\documentclass[preprint,3p,12pt]{elsarticle}

\usepackage{amssymb}
\usepackage{amsmath}
\usepackage{graphicx}%
\usepackage{multirow}%
\usepackage{amsmath,amssymb,amsfonts}%
\usepackage{amsthm}%
\usepackage{mathrsfs}%
\usepackage[title]{appendix}%
\usepackage{xcolor}%
\usepackage{textcomp}%
\usepackage{manyfoot}%
\usepackage{booktabs}%
\usepackage{algorithm}%
\usepackage{algorithmicx}%
\usepackage{algpseudocode}%
\usepackage{listings}%
\usepackage[caption=false,font=footnotesize]{subfig}
\usepackage{threeparttable}
\usepackage{float}

\theoremstyle{definition}
\newtheorem{definition1}{Analysis}

\journal{Engineering Applications of Artificial Intelligence}

\makeatletter
\def\ps@pprintTitle{%
  \let\@oddhead\@empty
  \let\@evenhead\@empty
  \let\@oddfoot\@empty
  \let\@evenfoot\@empty}
\makeatother

\begin{document}

\begin{frontmatter}



\title{Redefining Clustered Federated Learning for System Identification: The Path of ClusterCraft}

\author[itu]{Ertu\u{g}rul Ke\c{c}eci\corref{cor1}}
\ead{kececie@itu.edu.tr}

\author[itu]{M\"{u}jde G\"{u}zelkaya}
\ead{guzelkaya@itu.edu.tr}

\author[ai]{Tufan Kumbasar}
\ead{kumbasart@itu.edu.tr}

%% Affiliations
\affiliation[itu]{
    organization={Faculty of Electrical and Electronics Engineering},
       city={Istanbul},
            postcode={34469},
            country={Türkiye}
}

\affiliation[ai]{
    organization={Artificial Intelligence and Intelligent Systems Laboratory},
         city={Istanbul},
            postcode={34469},
            country={Türkiye}
}

%% Corresponding Author
\cortext[cor1]{Corresponding author}

%% Abstract
\begin{abstract}
This paper addresses the System Identification (SYSID) problem within the framework of federated learning. We introduce a novel algorithm, Incremental Clustering-based federated learning method for SYSID (IC-SYSID), designed to tackle SYSID challenges across multiple data sources without prior knowledge. IC-SYSID utilizes an incremental clustering method, ClusterCraft (CC), to eliminate the dependency on the prior knowledge of the dataset. CC starts with a single cluster model and assigns similar local workers to the same clusters by dynamically increasing the number of clusters. To reduce the number of clusters generated by CC, we introduce ClusterMerge, where similar cluster models are merged. We also introduce enhanced ClusterCraft to reduce the generation of similar cluster models during the training. Moreover, IC-SYSID addresses cluster model instability by integrating a regularization term into the loss function and initializing cluster models with scaled Glorot initialization. It also utilizes a mini-batch deep learning approach to manage large SYSID datasets during local training. Through the experiments conducted on a real-world representing SYSID problem, where a fleet of vehicles collaboratively learns vehicle dynamics, we show that IC-SYSID achieves a high SYSID performance while preventing the learning of unstable clusters. 
\end{abstract}


% begin{highlights}
% \item Proposes an incremental clustering-based FL method for system identification.\item Introduces ClusterCraft for incremental clustering without prior knowledge of data.
% \item Develops ClusterMerge and enhanced ClusterCraft to optimize clustering.
% \item Validates IC-SYSID on a real-world vehicle dataset, achieving high SYSID performance.
% \end{highlights}\

%% Keywords

\end{frontmatter}




\section{Introduction}

System Identification (SYSID) aims to estimate a model for system dynamics using observed input-output data, enabling accurate prediction and simulation of system behavior \cite{ljung2010perspectives,keesman2011system}. Linear-time Invariant (LTI) SYSID focuses on estimating the linear models while nonlinear SYSID captures complex system dynamics employing different methods such as nonlinear state-space models \cite{schon2011system}, Koopman operator \cite{bruder2019nonlinear}, and neural networks \cite{chiuso2019system, tuna2022deep}. Yet, existing studies typically consider a single data source, overlooking the potential benefits of utilizing multiple data sources from systems with similar dynamics, which can enhance sample efficiency. \cite{xin2022identifying, zhang2023multi, wang2023fedsysid, papusha2014collaborative}. 

Federated Learning (FL) is a machine learning approach where a shared global model in a center server is trained by aggregating locally trained models from decentralized clients (local workers) without sharing their data \cite{mcmahan2017communication, le2021federated, kaheni2024selective}. However, FL performs deficiently when the data among the local workers are heterogeneous and non-iid \cite{zhao2018federated}. Several studies have addressed this problem of FL by modifying local training or global model aggregation \cite{li2020federated, li2023fedlga, yeganeh2020inverse, zhang2022fine}. Moreover, another idea to handle the heterogeneous and non-iid data is to represent the diversity among local workers with more than one global model, i.e. cluster models \cite{ghosh2020efficient, mansour2020three, li2021federated, ruan2022fedsoft}.

The first implementation of FL into the SYSID problem is proposed in \cite{wang2023fedsysid}, where multiple systems are considered as the local workers collaboratively contributing to the learning of a global model that represents their dynamics. Nevertheless, performing SYSID for multiple systems, as in \cite{wang2023fedsysid} does not allow the learning of personalized models, thus, SYSID performance is affected by system heterogeneity. To address this problem, the clustered FL approach, Iterative Federated Clustering Algorithm (IFCA) \cite{ghosh2020efficient}, is implemented to the SYSID problem in Clustered SYSID (C-SYSID) algorithm \cite{toso2023learning} where system heterogeneity is represented by learning multiple cluster models. It is shown that C-SYSID theoretically guarantees convergence and performs efficiently under certain assumptions. Within this paper, we underline the following main issues of C-SYSID through investigations: (1) stability of cluster models, (2) training large-scale data, and (3) dependency on prior knowledge of the SYSID dataset. This work builds on our preliminary study in \cite{kececi2024novel} by conducting a deeper investigation into the SYSID issues of C-SYSID, extending IC-SYSID with improved clustering and a new evaluation on a realistic vehicle dataset.

In this paper, we redefine clustered FL by proposing an Incremental Clustering-based FL method for SYSID (IC-SYSID) that is tailored for SYSID challenges arising from multiple data sources. 
\begin{itemize}
    \item \textit{To learn stable cluster models}, we add a regularization term into the loss function of each local worker. Moreover, we suggest a scaled version of Glorot initialization to initialize stable cluster models for prioritizing stability from the outset. 
    \item \textit{To tackle the training of large-scale datasets}, we define a deep learning framework for the training of local worker models. This framework integrates mini-batched deep learning optimizers and includes a validation phase to enhance training performance.  
    \item \textit{To eliminate the dependency on prior knowledge of the multiple source dataset}, we propose an incremental clustering method: ClusterCraft (CC). CC initiates with a single cluster and assigns similar local workers to the same clusters while conditionally increasing the number of clusters if needed. To reduce the number of clusters generated by CC, we introduce ClusterMerge (CM) where similar cluster models are merged. As CM occurs after the training, we also develop an enhanced version of CC (eCC) to generate fewer similar clusters during the training phase to further optimize the clustering and SYSID performance.
\end{itemize}

The performance of IC-SYSID is evaluated on a realistic scenario in which a fleet of vehicles collectively learns cluster models representing their dynamics without prior knowledge of vehicle dynamics feasibility. Through comparative analyses, we demonstrate that IC-SYSID performs remarkably better than C-SYSID in both prediction and simulation modes. We show that IC-SYSID extensively reduces the learning of unstable cluster models while eCC reduces the number of communication rounds and generates fewer clusters when compared to its CC and CC-CM counterparts.

The paper is structured as follows: Section \ref{prelim} provides background on SYSID and clustered FL alongside the description of a dataset that serves as a running example. Section \ref{Csysid} covers C-SYSID, while Section \ref{incsysid} presents the proposed IC-SYSID. A comparative case study is detailed in Section \ref{exp}, followed by the driven conclusions and future work in Section \ref{conc}. Table \ref{notations} lists the commonly used notations throughout the paper.

\begin{table}[t]
\centering
\footnotesize
\renewcommand{\arraystretch}{1.3}
\setlength{\tabcolsep}{5pt}
\caption{Widely Used Notations}
\label{notations}
\begin{threeparttable}
\begin{tabular}{@{}ll@{}}
\toprule
\midrule
\( R \) & number of communication rounds \\ 
\( M \) & number of local workers \\ 
\( K \) & number of clusters \\
\( Q \) & number of ground-truth clusters \\ 
\( W^i \) & \(i^{\text{th}}\) local worker, \(i=1, \ldots, M\) \\ 
\( C^j \) & \(j^{\text{th}}\) cluster label, \(j=1, \ldots, K\) \\ 
\( C^q \) & \(q^{\text{th}}\) ground-truth cluster label, \(q=1, \ldots, Q\) \\ 
\( \theta^{(q)} \) & model parameters of \(C^q\) \\ 
\( \tilde{\theta}^{(j)} \) & estimated model parameters of \(C^j\) \\ 
\( \boldsymbol{\theta} \) & set of estimated model parameters \\ 
\( \hat{j} \) & cluster identity \\ 
\( \mathbf{F}_r \) & cluster identity flag, \(r=0, \ldots, R\) \\ 
\( \underline{P}^{(i)} \) & performance metric of \(W^i\) \\ 
\( \bar{P}^{(q)}_n \) & performance metric of \(C^q\) \\ \bottomrule
\end{tabular}
\end{threeparttable}
\end{table}

\section{Preliminaries} \label{prelim}

\subsection{System Identification}
Consider the following nonlinear system 
\begin{equation}
\label{nlti}
    \boldsymbol{x}_{t+1} = f(\boldsymbol{x}_t,\boldsymbol{u}_t,\boldsymbol{w}_t), \quad t = 1,2, ..., T
\end{equation}
where $\boldsymbol{x}_t = (x_{1,t}, \dots, x_{N_x,t})^T$ represents the states, $\boldsymbol{u}_t = (u_{1,t}, \dots, u_{N_u,t})^T$ the inputs, and $\boldsymbol{w}_t = (w_{1,t}, \dots, w_{N_x,t})^T$ the noise while $f(\cdot)$ encapsulates the nonlinearity. The corresponding SYSID problem with a Dataset $D=\{\boldsymbol{x},\boldsymbol{u}\}$ is solved by minimizing the following loss:
\begin{equation}
\label{rawloss}
    \mathcal{L} = \frac{1}{T}\sum_{t=1}^{T}\|\boldsymbol{x}_t - \tilde{\boldsymbol{x}}_t\|_2^2
\end{equation}
where $\boldsymbol{x}_t$ is the measured states while $\tilde{\boldsymbol{x}}_{t}$ is the estimated ones. If it is desired to approximate $f(\cdot)$ with an LTI model, $\tilde{\boldsymbol{x}}_{t}$ is generated via: 
\begin{equation} \label{LTI}
    \tilde{\boldsymbol{x}}_{t+1} = \tilde{A}\boldsymbol{x}_{t} + \tilde{B}\boldsymbol{u}_{t}, \quad t = 1,2, ..., T   
\end{equation}
Here, $\tilde{A} \in \mathbb{R}^{N_x \times N_x}$, and $\tilde{B} \in \mathbb{R}^{N_x \times N_u}$ are the state and input matrices of the model, respectively. To transform the SYSID problem into a regression problem, we define a regressor dataset  $\Phi = [\boldsymbol{x}_{1:T};\ \boldsymbol{u}_{1:T}] \in \mathbb{R}^{(N_x+N_u) \times T}$ with a target $X = [\boldsymbol{x}_{2:T+1}] \in \mathbb{R}^{N_x \times T}$. Then, the learnable parameter set is $\tilde{\theta} = [\tilde{A} \ \tilde{B}] \in \mathbb{R}^{N_x \times (N_x+N_u)}$.

\subsection{Clustered Federated Learning} \label{pfl}

In FL, there are $M$ decentralized clients which are local workers, $W^1, W^2 \dots, W^M$,  with distinct datasets  $\boldsymbol{{D}}=\{D^1, D^2 \dots, D^M\}$ and one center server that defines the global model. In each communication round $(r=0, \dots, R-1)$, the central server broadcasts global model parameters $\tilde{\theta}_r$ to all local workers. Then, each $W^i$ individually trains their local model parameters $\tilde{\theta}^{(i)}$ using $D^i$ and sends $\tilde{\theta}^{(i)}$ back to the central server. The central server aggregates $\tilde{\theta}^{(i)}$ to update $\tilde{\theta}$. A single global model aggregation is problematic when the data distribution of $\boldsymbol{D}$ is not normal or multimodal. This issue can be solved by defining more than one global model, i.e., cluster models \cite{ghosh2020efficient, mansour2020three}. 

In IFCA, it is assumed that $M$ local workers belong to $K$ clusters labeled as $C^1, C^2 \dots, C^K$, rather than a single one \cite{ghosh2020efficient}. In the communication round, each  $W^i$ defines its cluster identity $\hat{j}$ by finding $C^{\hat{j}}$ as follows: 
\begin{equation}
\label{pflidentity}
    \hat{j} = \text{argmin}_{j\in[K]}\mathcal{L}(\tilde{\theta}^{(j)})
\end{equation}
Then, each $W^i$ trains $\tilde{\theta}^{(\hat{j})}$ and sends back $\tilde{\theta}^{(i)}$ to the center server. The center server aggregates $\tilde{\theta}^{(i)}$ by taking account $\hat{j}$. In this context, a one-hot encoding vector $s_i$ is used to encode the cluster identity of $W^i$ to each cluster as follows: 
\begin{equation}
\label{onehotpfl}
    s_i = (s_{i,j})_{j=1}^{K} \ \text{with} \ s_{i,j} = \mathbf{1}\{j = \hat{j}\}
\end{equation}
where $s_{i, j}$ refers to cluster identity of $W^i$ to $C^j$. Using $s_i$, each $C^j$ is updated by the center server as follows
\begin{equation}
\label{fedavg}
    \tilde{\theta}^{(j)}=\sum_{i \in M} s_{i, j} \tilde{\theta}^{(i)} / \sum_{i \in M} s_{i, j}, \forall j \in [K]    
\end{equation}
where $\tilde{\theta}^{(j)}$ denotes the model parameters of $C^j$ whereas $\tilde{\theta}^{(i)}$ is the model parameters of $W^i$.

\subsection{Synthetic dataset} \label{synt}
As a running example, to evaluate the SYSID performances, we generate a synthetic dataset akin to \cite{toso2023learning}. We gather a dataset ($\boldsymbol{D}$) for $M$ local workers via the following LTI models: 
\begin{equation}
\label{ltiwithnoise}
    \boldsymbol{x}_{t+1}^{(i)} = A^{(i)}\boldsymbol{x}_t^{(i)} + B^{(i)}\boldsymbol{u}_t^{(i)} + \boldsymbol{w}_t^{(i)}, \quad t=1,2, \dots, T
\end{equation}
where the initial states, inputs, and noises are distributed by $\boldsymbol{x}_1^{(i)} \sim \mathcal{N}(0, \sigma_{x}^2 I_{N_x})$, $\boldsymbol{u}_{1:T}^{(i)} \sim \mathcal{N}(0, \sigma_{u}^2 I_{N_u})$ and $\boldsymbol{w}_{1:T}^{(i)} \sim \mathcal{N}(0, \sigma_{w}^2 I_{N_x})$, respectively.  
We have access to $M$ datasets corresponding to observed system trajectories and each of them is generated by one of the $Q$ different ground-truth clusters ($C^1, C^2 \dots, C^Q$) defined with model parameters $\theta^{(q)} = [A^{(q)} \ B^{(q)}]$. For the cluster $C^q$, we set $A^{(i)} = A^{(q)}, \ B^{(i)} = B^{(q)}, \ \forall i \in C^q$ to generate datasets. We refer a single trajectory $\{\boldsymbol{x}_t^{(i)}, \boldsymbol{u}_t^{(i)}\}$ as rollout for $W^i \in C^q$. We collect $N$ rollouts where each rollout consists of system trajectories as $\{\boldsymbol{x}_{t}^{(i)}, \boldsymbol{u}_{t}^{(i)}\}_{t=1}^T$.

We define $Q=5$ ground-truth clusters and generate distinct datasets for $M=100$ local workers. The number of systems for each $C^q$ is $|{C}^1| = 10, |{C}^2| = 24, |{C}^3| = 16, |{C}^4| = 28$ and $|{C}^5| = 22$. The parameters of each cluster model are given in the Appendix. For each local worker, we collected $T=50$ samples for $N=100$ rollouts with $\sigma_{x} = \sigma_{u} = 0.05$ and $\sigma_w = 0.02$.

\section{Clustered SYSID: Background and Drawbacks} \label{Csysid}
C-SYSID \cite{toso2023learning} is a straightforward deployment of IFCA into the area of SYSID. In this preliminary study, it is assumed that $\boldsymbol{D}$ is generated from $K$ different clusters, and each cluster is defined with an LTI model. The performance of C-SYSID is measured through its parameter estimation performance:    
\begin{equation}
\label{csiest}
    e_R^{(j)} = ||\tilde{\theta}^{(j)} - \theta^{(j)}||_F, \ \forall j\in [K]
\end{equation}
where $||\cdot||_F$ is the Frobenius norm. The authors assumed to know $\theta^{(j)}$, i.e, the ground-truth model parameters of $C^j$.

In this section, we discuss the main issues of C-SYSID and provide analyses using the implementation provided in \cite{toso2023learning} on the Synthetic dataset. All experiments are repeated 10 times with the hyperparameters given in the Appendix.

\subsection{Issue-1: Stability}

The stability of the learned cluster model in \eqref{LTI} depends obviously on eigenvalues ($\lambda$) of the state matrix $\tilde{A} \in \tilde{\theta}^{(j)}$. For stability, it is well-known that $|\lambda| < 1, \ \forall \lambda \in \boldsymbol{\Lambda}$. 

In the loss $\mathcal{L}$ of C-SYSID given in \eqref{rawloss}, there is no constraint or a regularization term to enforce that the learned cluster models will result in stable dynamics. Thus, a local worker $W^i$ might learn an unstable $\tilde{A}^{(i)} \in \tilde{\theta}^{(i)}$, i.e, $|\lambda| \geq  1, \ \exists \lambda \in \boldsymbol{\Lambda}$. As a consequence, a single unstable $\tilde{A}^{(i)}$ may yield an unstable  $\tilde{A}^{(j)}$ as $\tilde{\theta}^{(j)}$ is calculated
using \eqref{fedavg}.

\subsection{Issue-2: Training of a Local Worker}

In C-SYSID, the well-known least squares update rule is deployed to learn $\tilde{\theta}^{(i)}$ of each local worker. 

They created a target and the regressor for a single rollout $l$ as $X_l^{(i)} = [\boldsymbol{x}_{l, T+1:2}^{(i)}]$ and ${\Phi}_l^{(i)} = [\boldsymbol{x}_{l, T:1}^{(i)}; \ \boldsymbol{u}_{l, T:1}^{(i)} ]$, respectively. The target and the regressor of a $W^i \in C^q$ is defined as $X^{(i)} = [X^{(i)}_1, \dots, X^{(i)}_N] \in \mathbb{R}^{N_x \times N T}$ and $\Phi^{(i)} = [\Phi^{(i)}_1, \dots, \Phi^{(i)}_N] \in \mathbb{R}^{N_u \times N T}$. The update rule for $W^i$ is as follows:

\begin{equation}
\label{ls}
    \tilde{\theta}_{r+1}^{(i)} = \tilde{\theta}_{r}^{(i)} + \alpha (X^{(i)} - \tilde{\theta}_{r}^{(i)} \Phi_{r}^{(i)})\Phi^{(i)^T}_{r}
\end{equation}
where $\alpha$ is the learning rate. 

The issue with using the least square estimation is that if the dataset is large ($T\gg N_x+N_u$), the training cost is increased significantly. Besides, each worker has to allocate its entire data for updates, which increases the memory usage.

\subsection{Issue-3: Dependency on Prior Knowledge of Dataset}

In C-SYSID, it is assumed that we have prior knowledge of the cluster models in the dataset $\boldsymbol{D}$ which is used to set not only the number of cluster models $K$ but also to initialize the cluster model parameters $\tilde{\theta}^{(j)}$. It has been stated in \cite{toso2023learning} that the initial $\tilde{\theta}^{(j)}$ satisfy the following condition
\begin{equation}
\label{warm}
    ||\tilde{\theta}^{(j)} - \theta^{(j)}||_2 < (\frac{1}{2} - \eta)\Delta_{min} , \ \forall j \in [K], \ 0 < \eta < \frac{1}{2}
\end{equation}
where $\Delta_{min} = \min_{j\neq j'} ||\theta^{(j)} - \theta^{(j')}||_2$ is the minimum separation between model parameters of ground-truth clusters. However, the initialization of $\tilde{\theta}^{(j)}$ satisfying \eqref{warm} requires prior knowledge of ground truth cluster model parameters ${\theta}^{(j)}$.

In SYSID problems, the accessibility of such rich prior knowledge is not always feasible, as it requires analyzing the $\boldsymbol{D}$, making the initialization unusable and leaving the setting of $K$ undetermined. Thus, we investigated two aspects: (1) the performance of C-SYSID with random initialization and (2) the impact of the $K$ setting on its performance.

\begin{definition1} \label{Init}
Let us assume that the C-SYSID is facing a dataset $\boldsymbol{D}$ with no prior knowledge and consisting of $Q (Q \neq K, Q \le M)$ different clusters. Thus, as $Q \neq K$, we reformulated the metric in \eqref{csiest} as follows:
\begin{equation}
\label{est}
    e_R^{(j,q)} = ||\tilde{\theta}^{(j)} - \theta^{(q)}||_F, \ \forall j\in [K], \forall q\in [Q]
\end{equation}
where $\theta^{(q)}$ is the ground truth model parameters of $q^\text{{th}}$ cluster. Now, since there will be $K$ different $e_R^{(j,q)}$ for  ${\theta}^{(q)}$, we assign: 
% the minimum of $e_R^{(j,q)}$,
\begin{equation}
\label{redest}
    e_R^{(q)} = \min_j e_R^{(j,q)}, \ \forall q\in [Q]
\end{equation}
as the estimation measure for C-SYSID. Note that, if $j=q$ then \eqref{est} is identical to \eqref{csiest}.

We first set $K=5$ (i.e., $K=Q$) and analyzed the performance of C-SYSID using Glorot initialization, which is a widely used one, in comparison with the warm one. Then, we investigated its performance for $K = \{2,10\}$ (i.e. $K\neq Q$) where $\tilde{\theta}^{(j)}$ are initialized with Glorot. In Table \ref{tab:prob2}, we provided the mean $e_R^{(q)}$ to analyze the SYSID performance. Moreover, to investigate the performance of the cluster identity mechanism given in \eqref{pflidentity}, we present the Misclassified local Worker percentage (MW\%). We define a local worker as misclassified if it trains the same $\tilde{\theta}^{(j)}$ as another local worker belonging to a different $C^q$. Observe that:
\begin{itemize} 
    \item Warm initialization of $\tilde{\theta}^{(j)}$ with $K=5$ ensures effective operation of the cluster identity evaluation given in \eqref{pflidentity}. With prior knowledge aligning $\tilde{\theta}^{(j)}$ with ground-truth clusters, correct identities are assigned to local workers, resulting in zero MW\% and satisfactory performance.
    \item Random initialization of $\tilde{\theta}^{(j)}$ regardless the setting of $K$ results in a poor SYSID performance. The cluster identity mechanism incorrectly assigns identical cluster identities to local workers from different ground-truth clusters, leading to collective contribution to a single $\tilde{\theta}^{(j)}$. Thus, in subsequent rounds, it continues to assign dissimilar workers to a single cluster model for updates, as indicated by the high MW\%.
\end{itemize}

To sum up, the cluster identity mechanism of C-SYSID is highly sensitive to the initialization of $\tilde{\theta}^{(j)}$ and configuration of $K$, which are huge drawbacks.  
\end{definition1}

\begin{table}[t]
\centering
\footnotesize
\renewcommand{\arraystretch}{1.3}
\setlength{\tabcolsep}{4pt}
\caption{Performance of C-SYSID over 10 experiments}
\label{tab:prob2}
\begin{threeparttable}
\begin{tabular}{@{}lccccccr@{}}
\toprule
\textbf{K} 
& \textbf{Initialization} 
& \(\mathbf{e_{R}^{(1)}}\) 
& \(\mathbf{e_{R}^{(2)}}\) 
& \(\mathbf{e_{R}^{(3)}}\) 
& \(\mathbf{e_{R}^{(4)}}\) 
& \(\mathbf{e_{R}^{(5)}}\) 
& \textbf{MW\%}
\\ \midrule

5 
& Warm \cite{toso2023learning} 
& 0.016 
& 0.004 
& 0.011 
& 0.004 
& 0.005 
& 0\%
\\

5 
& Glorot 
& 1.182 
& 0.904 
& 0.918 
& 0.503 
& 0.798 
& 41.2\%
\\ \midrule

2 
& Glorot 
& 1.324 
& 1.241 
& 0.984 
& 0.760 
& 1.454 
& 33.8\%
\\

10 
& Glorot 
& 1.117 
& 0.415 
& 0.820 
& 0.254 
& 0.316 
& 20.8\%
\\ \bottomrule

\end{tabular}
\end{threeparttable}
\end{table}

\section{Incremental Clustering based FL for SYSID} \label{incsysid}

In this section, we introduce IC-SYSID and show how it addresses each problem of C-SYSID. The training flow of IC-SYSID is outlined in Algorithm \ref{alg:alg1}\footnote{A GitHub repository will be shared upon acceptance of the paper}. 

We evaluate its performance through analyses derived from experiments conducted on the Synthetic dataset. All experiments are repeated 10 times with the hyperparameters provided in the Appendix.

\begin{algorithm}
\caption{IC-SYSID}\label{alg:alg1}
    \begin{algorithmic}[1]
        % \small
        \State \textbf{Initialization:} number of communication rounds $R$, number of clusters $K_{0}=1$, initialization of $\tilde{\theta}^{(j)}_{0}, j \in [K_{0}]$ using \eqref{glorot}, initial cluster flag $\textbf{F}_0 = \mathbf{1} \in \mathbb{R}^{M}$
        \For{$r=0,1,2, \dots, R-1$}
            \State \textsc{\underline{Center Server}:}
            \For{$i=1,2, \dots, M$}
                \State $\hat{j}=\mathbf{F}_r^{(i)}$
                \State broadcast $\tilde{\theta}^{(\hat{j})}_{r}$
            \EndFor
            \State \textsc{\underline{Worker Client}:}
            \For{each local worker $W^i$ \textbf{in parallel}}
                \State $s_i = (s_{i,j})_{j=1}^{K_{r}}$ with $s_{i,j} = \mathbf{1}\{j = \hat{j}\}$
                \State $\tilde{\theta}_{r+1}^{(i)}$ $\gets$ \textit{LocalUpdate}$(\tilde{\theta}_{r}^{(i)}, \alpha)$
                \State calculate $P^{(i)}$ and $\Bar{\mathcal{L}}^{(i)}$ using $D^i_{CC}$
                \State send $s_i$, $P^{(i)}$, $\Bar{\mathcal{L}}^{(i)}$ and $\tilde{\theta}_r^{(i)}$ to center server
            \EndFor
            \State \textsc{\underline{Center Server}:}
            \State \textbf{option I} (ClusterCraft):
            \State \quad $K_{r+1}, \textbf{F}_{r+1} $ = \textit{{CC}}$(\underline{P}_{r-1},P_r, K_r, \textbf{F}_r)$
            \State \textbf{option II} (Enhanced ClusterCraft):
            \State \quad $K_{r+1}, \textbf{F}_{r+1} $ = \textit{{eCC}}$(\underline{P}_{r-1},P_r, K_r, \textbf{F}_r)$ 
            \vspace{0.05cm}
            \State $\tilde{\theta}^{(j)}_{r+1} = \sum_{i \in M} \tilde{\theta}^{(i)}_{r+1} / \sum_{i \in M} s_{i,j} $, $\forall j \in [K_{r}]$
            \vspace{0.05cm}
            \If{$\Bar{\mathcal{L}}^{(i)} < \epsilon_l, \ \forall i \in [M]$}
            \State \textbf{goto} \textbf{end for}
            \EndIf
        \EndFor
        \State \textbf{option} (ClusterMerge):
        \State \quad \textbf{Return} $\boldsymbol{\theta} = \textit{ClusterMerge}(\boldsymbol{\theta})$
        \State \textbf{Return} $\boldsymbol{\theta} = \{\tilde{\theta}^{(1)}, \tilde{\theta}^{(2)}, \dots, \tilde{\theta}^{(K_R)}\}$
        \vspace{0.2cm}
        \State \textit{\underline{LocalUpdate}}$(\tilde{\theta}^{(i)}, \alpha)$ 
        \State \textbf{Initialization:} $D^i = \{D^{i}_T, D^{i}_{CC}\}$, mini-batch size $mbs$, learning rate $\alpha$, split $D^i_T$ into mini-batches of size $mbs$, Lagrange multiplier $\mu$, parameters of Adam optimizer $\beta_1$ and $\beta_2$
        \State $m=0, \ v=0$
        \For{each epoch}
            \For{each mini-batch}
            \State $\tilde{\theta}^{(i)} \gets$ AdamUpdate$(\tilde{\theta}^{(i)}, m, v, \alpha)$
            \EndFor
        \EndFor
        \State \textbf{Return} $ \tilde{\theta}_{r+1}^{(i)} \gets \tilde{\theta}^{(i)}$
    \end{algorithmic}
\end{algorithm}

\subsection{Stability of Local Worker}

In IC-SYSID, we aim to learn local model parameters $\tilde{\theta}^{(i)}$ that are inherently stable to prevent instability during the aggregation via  \eqref{fedavg}. To address this, we incorporate a regularization term into $\mathcal{L}$ of each local worker as follows: 

\begin{equation}
\label{loss}
\begin{gathered}
    \mathcal{L} = \frac{1}{T}\sum_{t=1}^{T}\|\boldsymbol{x}_t - \tilde{\boldsymbol{x}}_t\|_2^2 + \mu ||\tilde{A}||_F    
\end{gathered}
\end{equation}
The regularization term $||\tilde{A}||_F$ aims to minimize the magnitude of each element of $\tilde{A}$ to implicitly enforce $|\lambda| < 1, \ \forall \lambda \in \boldsymbol{\Lambda}$ and thus learn stable local models to be aggregated. It is crucial to note that optimizing \eqref{loss} does not strictly guarantee the stability of the cluster model. Yet, it is computationally efficient and easy to implement. Moreover, to ensure that cluster models are initially stable, we suggest the following modified Glorot initialization:  
\begin{equation}
\label{glorot}
\tilde{\theta}^{(j)} \sim U\left[-0.1\frac{\sqrt{6}}{\sqrt{N_x+N_u}}, 0.1\frac{\sqrt{6}}{\sqrt{N_x+N_u}}\right], \forall j\in [K]
\end{equation}
Note that we utilize a scaling factor of 0.1 to prevent the generation of initially unstable $\tilde{A}^{(j)}$, i.e, $|\lambda| \geq  1, \ \exists \lambda \in \boldsymbol{\Lambda}$.

\subsection{Local Worker Training}

In the IC-SYSID, rather than using the data of each worker $D^i$ completely for training, $D^i$ is split into training and CC sets $D^{i}=\{D^{i}_T, D^{i}_{CC}\}$. Here, $D^{i}_{CC}$ is used for performance evaluation by CC.

For training, to tackle the issues of large-scale data, we deploy a mini-batched deep learning approach. Subsequently, we partition each $D^{i}_T$ into mini-batches and utilize the Adam optimizer that is defined as follows:  

\begin{equation}
\label{adam}
\tilde{\theta}^{(i)} = \tilde{\theta}^{(i)} - \frac{\alpha}{\sqrt{\tilde{v}^{(i)}+ 10^{-6}}} \tilde{m}^{(i)}
\end{equation}
where $\tilde{m}^{(i)}$ and $\tilde{v}^{(i)}$ calculated as
\begin{equation}
    \tilde{m}^{(i)} = \frac{\beta_1 m^{(i)} + (1 - \beta_1) \nabla \mathcal{L}(\tilde{\theta}^{(i)})}{1 - \beta_1}
\end{equation}
\begin{equation}
    \tilde{v}^{(i)} = \frac{\beta_2 v^{(i)} + (1 - \beta_2) (\nabla \mathcal{L}(\tilde{\theta}^{(i)}))^2}{1 - \beta_2}
\end{equation}
We also integrated a standard early stopping mechanism into IC-SYSID using $D^i_{CC}$. Each local worker evaluates the moving average of its loss value over 10 communication rounds,
\begin{equation}
    \Bar{\mathcal{L}}^{(i)} = \frac{1}{10} \sum_{b=0}^9 \mathcal{L}^{(i)}_{r-b}
\end{equation}
and the training stops if there is no improvement of $\Bar{\mathcal{L}}^{(i)}, \forall W^{i}$. 

After each communication round, CC assesses the performance of each $W^i$ through the resulting normalized fit metric based on $D^{i}_{CC}$: 
\begin{equation}
\label{nfm}
    P^{(i)}_n = 1 - \sum_{t=1}^T \sqrt{\frac{x_{n,t}^{(i)} - \tilde{x}_{n,t}^{(i)}}{x_{n,t}^{(i)} - \overline{x}_{n}^{(i)}}}, \ n=1, 2, \dots, N_x
\end{equation}
where $\bar{x}_{n}^{(i)}= ( 1 \backslash T)\sum_{t=1}^T x_{n,t}^{(i)}$. Note that a perfect model fit for $n^{th}$ state results with $P^{(i)}_n = 1$. The overall performance of $W^i$ is determined based on its poorest state estimation result:
\begin{equation}
    \underline{P}^{(i)} = \min_n P^{(i)}_n
\end{equation}
Note that evaluating the performance using $\mathcal{L}$ might be problematic since its scale can vary depending on dataset distribution,  as well as on $\mu$ and $||\tilde{A}||_F$.

\subsection{Incremental Clustering:  ClusterCraft} \label{CC}

To eliminate the dependency on prior knowledge of the datasets, we developed incremental clustering methods capable of assigning similar $W^i$ to the same cluster and dynamically incrementing the number of clusters when dissimilarities arise.

CC initiates with a single cluster ($K_0=1$) and determines whether the number of clusters for the next communication round ($K_r$) should be increased. The cluster identities are set by CC running in the center server, rather than within the local worker akin to C-SYSID. CC logs and modifies the cluster identities of $M$ local workers (i.e., $\hat{j}$'s of all $W^i$) within a cluster identity flag $\textbf{F}_r \in \mathbb{R}^M$. For the first communication round, we set $\textbf{F}_0 = \textbf{1}$ since there is only a single cluster.

\subsubsection{ClusterCraft}
As given in Algorithm \ref{alg:alg2}, CC uses $\underline{P}^{(i)}$ to decide whether a local worker $W^i$ converged to a poor model fit if the following condition holds. 
\begin{equation}
\label{inq}
    |{\underline{P}}_r^{(i)} - {\underline{P}}_{r-1}^{(i)}| < \epsilon_\Delta \ \& \ {\underline{P}}_r^{(i)} < \epsilon_{p}  
\end{equation}
where ${\underline{P}}_r^{(i)}$ is the performance metric at the $r^\text{{th}}$ communication round, while $\epsilon_\Delta$ and $\epsilon_{p}$ are convergence thresholds. 

CC evaluates each $W^i$ and updates the cluster identity of any worker that meets the criteria in \eqref{inq}:
\begin{equation}
\label{flagupdate}
    \mathbf{F}_{r+1}^{(i)} = {K}_r + 1
\end{equation}
If any worker's cluster identity is updated, CC increases the number of clusters 
\begin{equation}
\label{clusterinc}
    K_{r+1} = K_r + 1
\end{equation}
and defines a cluster label $C^{{K}_{r+1}}$, initializing it with \eqref{glorot}. 

\begin{algorithm}[h]
\caption{CC: ClusterCraft}\label{alg:alg2}
\begin{algorithmic}[1]
    \State \textbf{Input:} ${\underline{P}}_{r-1}, {\underline{P}}_r, {K}_{r}, \mathbf{F}_{r}$
    \State \textbf{Require:} convergence thresholds $\epsilon_{\Delta}$ and $\epsilon_{p}$
    \For{$i=1,2, \dots, M$}
    \If{$|{\underline{P}}_r^{(i)} - {\underline{P}}_{r-1}^{(i)}| < \epsilon_{\Delta} \ \And \ {\underline{P}}_r^{(i)} < \epsilon_{p}$}
    \State $\mathbf{F}_{r+1}^{(i)} = {K}_r + 1$
    \Else
    \State $\mathbf{F}_{r+1}^{(i)} =\mathbf{F}_{r}^{(i)}$
    \EndIf
    \EndFor
    \If{$\mathbf{F}_{r+1} \neq \mathbf{F}_{r}$}
    \State ${K}_{r+1} = {K}_r + 1$
    \EndIf
    \State \textbf{Return}  ${K}_{r+1}, \mathbf{F}_{r+1}$
\end{algorithmic}
\end{algorithm}

\begin{table}[H]
\centering
\footnotesize
\renewcommand{\arraystretch}{1.4}
\setlength{\tabcolsep}{3pt}
\caption{Comparative Performance Analysis of IC-SYSID over 10 experiments: Synthetic dataset}
\label{tab:comp1}
\begin{threeparttable}
\begin{tabular}{@{}llccc|ccc@{}}
\toprule
\multicolumn{2}{c|}{} 
& \multicolumn{3}{c|}{\textbf{C-SYSID}}
& \multicolumn{3}{c}{\textbf{IC-SYSID}}
\\ \midrule
\multicolumn{2}{c|}{} 
& K=2 
& K=5 
& K=10 
& CC
& CC-CM 
& eCC 
\\ \midrule

$\bar{P}^{(1)}_1$ 
& 
& \(0.674 (\pm 0.087)\)
& \(0.677 (\pm 0.035)\)
& \(0.692 (\pm 0.054)\)
& \(0.749 (\pm 0.000)\)
& \(0.749 (\pm 0.000)\)
& \(0.749 (\pm 0.000)\)
\\
$\bar{P}^{(1)}_2$ 
& 
& \(0.076 (\pm 0.360)\)
& \(0.341 (\pm 0.323)\)
& \(0.382 (\pm 0.344)\)
& \(0.635 (\pm 0.000)\)
& \(0.635 (\pm 0.000)\)
& \(0.635 (\pm 0.000)\)
\\
$\bar{P}^{(1)}_3$ 
& 
& \(0.450 (\pm 0.036)\)
& \(0.498 (\pm 0.113)\)
& \(0.589 (\pm 0.154)\)
& \(0.781 (\pm 0.000)\)
& \(0.781 (\pm 0.000)\)
& \(0.781 (\pm 0.000)\)
\\ \midrule

$\bar{P}^{(2)}_1$ 
& 
& \(0.055 (\pm 0.226)\)
& \(0.178 (\pm 0.370)\)
& \(0.252 (\pm 0.393)\)
& \(0.749 (\pm 0.010)\)
& \(0.753 (\pm 0.000)\)
& \(0.753 (\pm 0.000)\)
\\
$\bar{P}^{(2)}_2$ 
& 
& \(0.713 (\pm 0.024)\)
& \(0.757 (\pm 0.043)\)
& \(0.761 (\pm 0.045)\)
& \(0.853 (\pm 0.017)\)
& \(0.858 (\pm 0.000)\)
& \(0.858 (\pm 0.000)\)
\\
$\bar{P}^{(2)}_3$ 
& 
& \(0.744 (\pm 0.071)\)
& \(0.789 (\pm 0.020)\)
& \(0.790 (\pm 0.021)\)
& \(0.839 (\pm 0.007)\)
& \(0.846 (\pm 0.000)\)
& \(0.846 (\pm 0.000)\)
\\ \midrule

$\bar{P}^{(3)}_1$ 
& 
& \(0.232 (\pm 0.113)\)
& \(0.315 (\pm 0.045)\)
& \(0.338 (\pm 0.115)\)
& \(0.558 (\pm 0.003)\)
& \(0.559 (\pm 0.003)\)
& \(0.561 (\pm 0.000)\)
\\
$\bar{P}^{(3)}_2$ 
& 
& \(0.550 (\pm 0.147)\)
& \(0.649 (\pm 0.082)\)
& \(0.616 (\pm 0.132)\)
& \(0.744 (\pm 0.009)\)
& \(0.745 (\pm 0.001)\)
& \(0.745 (\pm 0.000)\)
\\
$\bar{P}^{(3)}_3$ 
& 
& \(0.321 (\pm 0.170)\)
& \(0.417 (\pm 0.154)\)
& \(0.359 (\pm 0.203)\)
& \(0.589 (\pm 0.000)\)
& \(0.593 (\pm 0.001)\)
& \(0.593 (\pm 0.000)\)
\\ \midrule

$\bar{P}^{(4)}_1$ 
& 
& \(0.826 (\pm 0.038)\)
& \(0.837 (\pm 0.042)\)
& \(0.845 (\pm 0.046)\)
& \(0.897 (\pm 0.000)\)
& \(0.897 (\pm 0.000)\)
& \(0.897 (\pm 0.000)\)
\\
$\bar{P}^{(4)}_2$ 
& 
& \(0.866 (\pm 0.027)\)
& \(0.882 (\pm 0.012)\)
& \(0.886 (\pm 0.024)\)
& \(0.915 (\pm 0.000)\)
& \(0.915 (\pm 0.000)\)
& \(0.915 (\pm 0.000)\)
\\
$\bar{P}^{(4)}_3$ 
& 
& \(0.857 (\pm 0.016)\)
& \(0.863 (\pm 0.013)\)
& \(0.869 (\pm 0.020)\)
& \(0.895 (\pm 0.000)\)
& \(0.895 (\pm 0.000)\)
& \(0.895 (\pm 0.000)\)
\\ \midrule

$\bar{P}^{(5)}_1$ 
& 
& \(0.642 (\pm 0.116)\)
& \(0.695 (\pm 0.014)\)
& \(0.705 (\pm 0.009)\)
& \(0.695 (\pm 0.000)\)
& \(0.695 (\pm 0.000)\)
& \(0.695 (\pm 0.000)\)
\\
$\bar{P}^{(5)}_2$ 
& 
& \(0.665 (\pm 0.105)\)
& \(0.712 (\pm 0.042)\)
& \(0.739 (\pm 0.027)\)
& \(0.748 (\pm 0.000)\)
& \(0.748 (\pm 0.000)\)
& \(0.748 (\pm 0.000)\)
\\
$\bar{P}^{(5)}_3$ 
& 
& \(0.193 (\pm 0.380)\)
& \(0.342 (\pm 0.238)\)
& \(0.487 (\pm 0.171)\)
& \(0.557 (\pm 0.000)\)
& \(0.557 (\pm 0.000)\)
& \(0.557 (\pm 0.000)\)
\\ \midrule

$R$ 
& 
& 1000
& 1000
& 1000
& 191.3
& 191.3
& 100.2
\\
$K_R$ 
& 
& 2
& 5
& 10
& 45.8
& 5
& 5
\\
MW\%
& 
& 41.2\%
& 33.8\%
& 20.8\%
& 1\%
& 1\%
& 0\%
\\ \bottomrule

\end{tabular}
\vspace{0.5em}

\begin{tablenotes}[flushleft]
\item (1) Standard deviations of $\bar{P}_n^{(q)}$ are given within parentheses.
\item (2) $R$ and $K_R$ values are hyperparameters for C-SYSID.
\end{tablenotes}

\end{threeparttable}
\end{table}

\begin{definition1} \label{analysis3}
To investigate the efficiency of CC, we measure the SYSID performance with mean $P^{(q)}_n (\bar{P}^{(q)}_n)$ for each ground truth cluster:
\begin{equation}
\label{gclusterperf}
    \bar{P}^{(q)}_n = \frac{1}{|C^q|} \sum_{i \in C^q} {P^{(i)}_n}, \ \forall q \in [Q]
\end{equation}
Note that we have not used the measure in \eqref{redest} as we have assumed that we have no prior knowledge of $\theta^{(q)}$.  We also recorded $R$ to show the advantage of early stopping, and $K_R$ to examine the incremental clustering efficiency alongside MW\%.

In Table \ref{tab:comp1}, we provided the mean performance measures of IC-SYSID with CC. We also provided the $\bar{P}^{(q)}_n$ measure of C-SYSID with $K = \{2, 5, 10\}$ utilizing Glorot initialization for comparison.  We can observe that IC-SYSID with CC achieves significantly better performance than C-SYSID while yielding a remarkable decrease of MW\% with fewer communication rounds. It is worth noting that the performance of IC-SYSID with CC is consistent, as evidenced by the smaller standard deviation values provided within parentheses in Table \ref{tab:comp1}. 

We can conclude that CC successfully increases the number of clusters and assigns cluster identities of similar workers to the same cluster. Yet, it generates more clusters than the number of ground-truth clusters, $K_R \gg Q$.
\end{definition1}

\subsubsection{ClusterCraft with ClusterMerge}

To reduce $K_R$ of CC, we introduce the ClusterMerge. As outlined in Algorithm \ref{alg:alg3}, CM checks the similarities of the cluster model parameters, $\boldsymbol{\theta} = \{\tilde{\theta}^{(1)}, \dots, \tilde{\theta}^{(K_R)}\}$, and then merges the similar clusters. 

CM compares the similarities between the parameters with the predetermined similarity threshold $\epsilon_{\theta}$. 
\begin{equation}
\label{simmodels}
    ||\tilde{\theta}^{(j')} - \tilde{\theta}^{(j)}||_2 < \epsilon_{\theta}, \ \forall j \neq j'
\end{equation} 
Each $\tilde{\theta}^{(j)}$ that fulfills the condition is grouped into the parameter set of similar clusters, $\boldsymbol{\theta_{s}}$ which are then averaged. Then, $\boldsymbol{\theta_{s}}$ is added into the set of merged parameters $\boldsymbol{\theta_{m}}$. Subsequently, the CM removes $\boldsymbol{\theta_{s}}$ from $\boldsymbol{\theta}$ and repeats the process until $\boldsymbol{\theta}$ becomes an empty set.

\begin{algorithm}[t]
\caption{CM: ClusterMerge} \label{alg:alg3}
\begin{algorithmic}[1]
\State \textbf{Input: } $\boldsymbol{\theta} = \{\tilde{\theta}^{(1)}, \tilde{\theta}^{(2)}, \dots, \tilde{\theta}^{(K_R)}\}$
\State \textbf{Require: } similarity threshold $\epsilon_{\theta}$
\While{$\boldsymbol{\theta} \neq \varnothing$ }
    \State $\tilde{\theta}^{(j')} \gets \text{the first element of } \boldsymbol{\theta}$
    \State $\boldsymbol{\theta_{s}} \gets \{ \tilde{\theta}^{(j')} \}$ \Comment{Parameters set of similar clusters}
    \For{each $\tilde{\theta}^{(j)}$ in $\boldsymbol{\theta}$ where $j \neq j'$}
        \If{$||\tilde{\theta}^{(j')} - \tilde{\theta}^{(j)}||_2 < \epsilon_{\theta}$}
            \State $\boldsymbol{\theta_{s}} \gets \{ \tilde{\theta}^{(j)} \}$
        \EndIf
    \EndFor
    \State $\boldsymbol{\theta_{m}} \gets  (1 \backslash |\boldsymbol{\theta_{s}}|) \sum_{\tilde{\theta}^{(j)} \in \boldsymbol{\boldsymbol{\theta_{s}}}} \tilde{\theta}^{(j)}$
    \State $\boldsymbol{\theta} \gets \boldsymbol{\theta} \setminus \boldsymbol{\theta_{s}}$ \Comment{Remove similar clusters}
\EndWhile
\State \textbf{Return} $\boldsymbol{\theta_{m}}$ 
\end{algorithmic}
\end{algorithm}

\begin{definition1}

To show the efficiency of CM, we applied CM to the trained IC-SYSID using CC. We observe from Table \ref{tab:comp1} that the integration of CM with CC reduces $K_R$ by $89.1\%$ without impacting its SYSID performance.
\end{definition1}

\subsubsection{Enhanced ClusterCraft} 

We have observed that CC-CM successfully reduces $K_R$. Yet, CM occurs after the training of IC-SYSID. Since creating similar clusters during training can increase the number of communication rounds, we developed an enhanced version of CC, namely eCC, to minimize the number of similar clusters created during training.

Algorithm \ref{alg:alg4} presents the cluster generation process of eCC. In eCC, the center server compares the similarities between $W^{\hat{i}}$, which is the first local worker that satisfies \eqref{inq}, and all local workers with a predefined similarity threshold $\epsilon_{s}$. 
\begin{equation}
\label{manhattan}
    ||\tilde{\theta}^{(\hat{i})} - \tilde{\theta}^{(i)}||_1 < \epsilon_s, \ \forall i \in [M]
\end{equation}
eCC changes cluster identities of similar clusters which are the local workers fulfilling this condition, as given in \eqref{flagupdate}. eCC increases cluster number using \eqref{clusterinc} and initiates a new cluster $C^{K_{r+1}}$, if the cluster identity flag is modified. By changing the cluster identities of similar local workers, eCC assigns the cluster identities of similar local workers to the same cluster.

\begin{algorithm}[t]
\caption{eCC: Enhanced ClusterCraft}\label{alg:alg4}
\begin{algorithmic}[1]
    \State \textbf{Input:} $\underline{P}_{r-1} ,\underline{P}_{r} , {K}_{r}, \mathbf{F}_{r}, \tilde{\theta}_r$
    \State \textbf{Require:} convergence thresholds $\epsilon_{\Delta}$, $\epsilon_{p}$ and  similarity threshold $\epsilon_{s}$
    \For{$i=1,2, \dots, M$}
    \If{$|{\underline{P}}_r^{(i)} - {\underline{P}}_{r-1}^{(i)}| < \epsilon_{\Delta} \ \And \ {\underline{P}}_r^{(i)} < \epsilon_{p}$}
    \State $\hat{i}= i$
    \State \textbf{goto} \textbf{end for}
    \EndIf
    \EndFor
    \For{$i=1,2,\dots, M$}
    \If{$||  \tilde{\theta}^{(\hat{i})} - \tilde{\theta}^{(i)}||_1 < \epsilon_{s}$}
    \State $\mathbf{F}_{r+1}^{(i)} = {K}_r + 1$
    \Else
    \State $\mathbf{F}_{r+1}^{(i)} =\mathbf{F}_{r}^{(i)}$
    \EndIf
    \EndFor
    \If{$\mathbf{F}_{r+1} \neq \mathbf{F}_{r}$}
    \State ${K}_{r+1} = {K}_r + 1$
    \EndIf
    \State \textbf{Return}  ${K}_{r+1}, \mathbf{F}_{r+1}$
\end{algorithmic}
\end{algorithm}

\begin{definition1}
The results in Table \ref{tab:comp1} show that eCC performs very closely to CC. However, the generation of similar clusters during the training of IC-SYSID using eCC is significantly diminished while MW\% is decreased to 0\%. This advancement in creating fewer similar clusters leads to $47.3\%$ decrease in $R$. To sum up, eCC remarkably reduces both $K_R$ and $R$ while maintaining a performance very similar to that of CC-CM.
\end{definition1}

\section{A Case study: SYSID with a Fleet of Vehicles} \label{exp}

We consider a scenario where a fleet of vehicles (i.e. $M$ local workers) collectively perform SYSID to represent the vehicle dynamics. We assume that there is no prior knowledge of vehicle dynamics, implying that $\theta^{(q)}$ and $Q$ are unknown.

\subsection{Design of Experiments}

\begin{figure}[t]
    \centering    \includegraphics[width=0.7\linewidth]{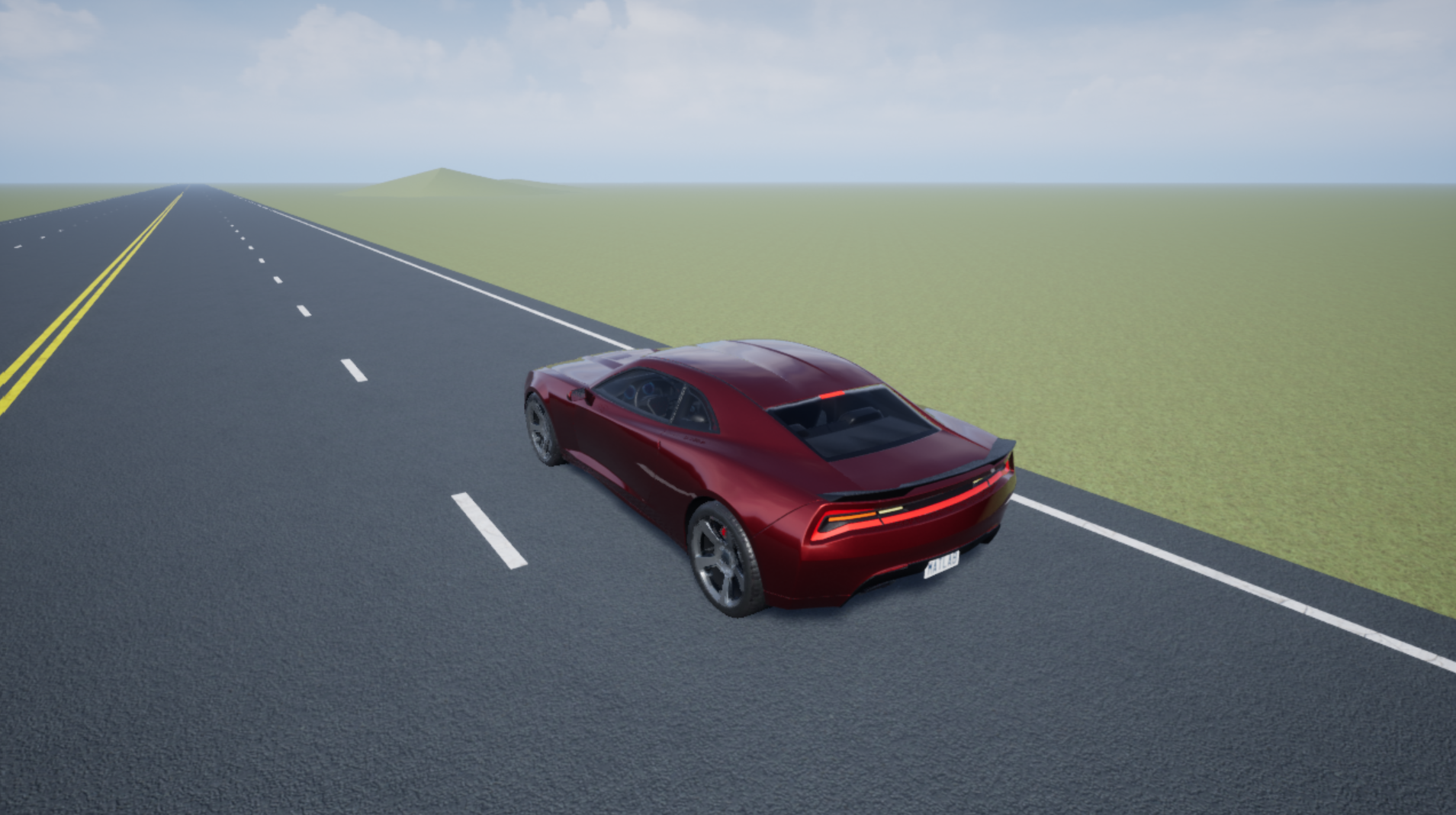}
    \caption{A snapshot of the realistic simulation environment}
    \label{fig:car}
\end{figure}

For realistic simulation, we used Matlab\textsuperscript{\textregistered} Vehicle Dynamics Blockset to simulate $M=50$ vehicles in high fidelity. A snapshot of the environment is given in Fig. \ref{fig:car}.

In the scope of SYSID, we only focused on modeling the lateral dynamics of the fleet vehicles under the assumption that the vehicles are cruising at constant steady-state longitudinal velocities $(v_{x})$. In this context, we employed a sinusoidal sweep signal as the steering angle reference $(\delta)$ with a magnitude of $45^\circ$ and swept it from $0$ Hz to $1$ Hz over 300 seconds. For each fleet vehicle, we defined the state to be estimated as $\boldsymbol{x} = [v_y; \dot{\psi}]$ through the input $\boldsymbol{u} = \delta$. To excite vehicle dynamics at different velocities, we defined 12 vehicles with $v_x=20$mph, 16 vehicles with $v_x=40$mph, 14 vehicles with $v_x=60$mph, and 8 vehicles with $v_x=80$mph. In each simulation, we gathered the steering angle as the input, $\boldsymbol{u} = \delta$, and the lateral velocity and yaw rate as the states, $\boldsymbol{x} = [v_y; \dot{\psi}]$. 

We trained and evaluated the IC-SYSID with CC, CC-CM, and eCC while C-SYSID with $K = \{2, 4, 8\}$. In C-SYSID, we initialized $\tilde{\theta}^{(j)}$ using Glorot initialization method as $\theta^{(q)}$ is unknown. All experiments were repeated with 10 different initial seeds with the hyperparameter configuration given in the Appendix. As a baseline, we utilized the Matlab\textsuperscript{\textregistered} System Identification Toolbox to estimate a State Space (SS) model for each vehicle. 

\subsection{Performance Measures}
We repeated all experiments 10 times and evaluated the SYSID performances of a single experiment with
\begin{equation}
\label{simcar}
    \Bar{P}_n = \frac{1}{M} \sum_{i \in [M]} P_n^{(i)}
\end{equation}
rather than \eqref{gclusterperf} as $Q$ is not known in advance. Given the importance of learning stable models in SYSID problems, we calculated the percentage of unstable cluster models relative to the total cluster models over 10 experiments (UC\%). We have also recorded the mean $R$ and $K_R$ to analyze the clustering performance of IC-SYSID.

\begin{figure*}[t]
    \centering
    \includegraphics[width=1\linewidth]{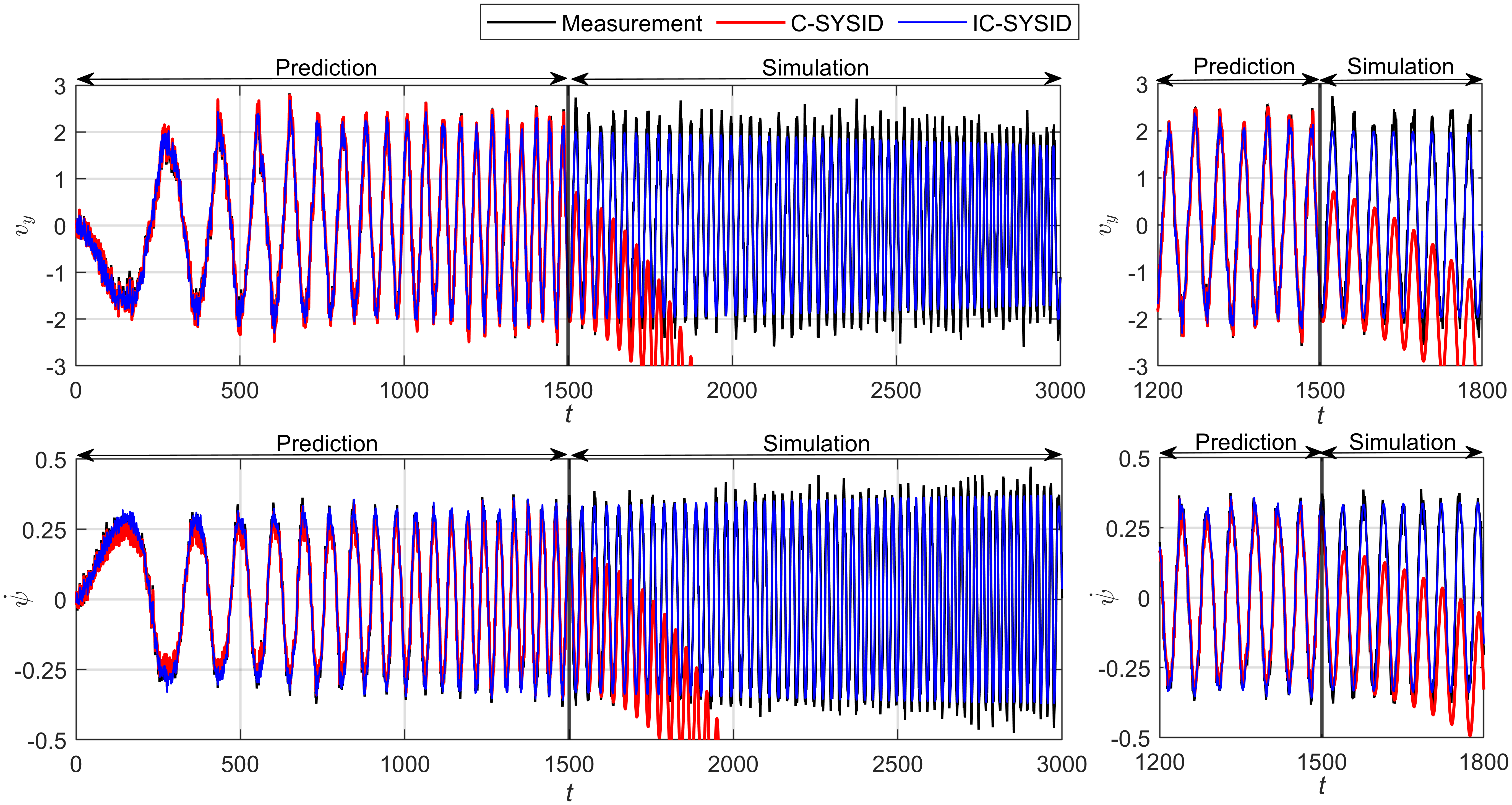}
    \caption{Comparison of C-SYSID with $K=2$ and IC-SYSID with CC on Car Dataset (full-scale plot on the left, zoomed-in plot on the right)}
    \label{fig:car-data-comp}
\end{figure*}

\begin{figure*}[t]
    \centering
    \includegraphics[width=1\linewidth]{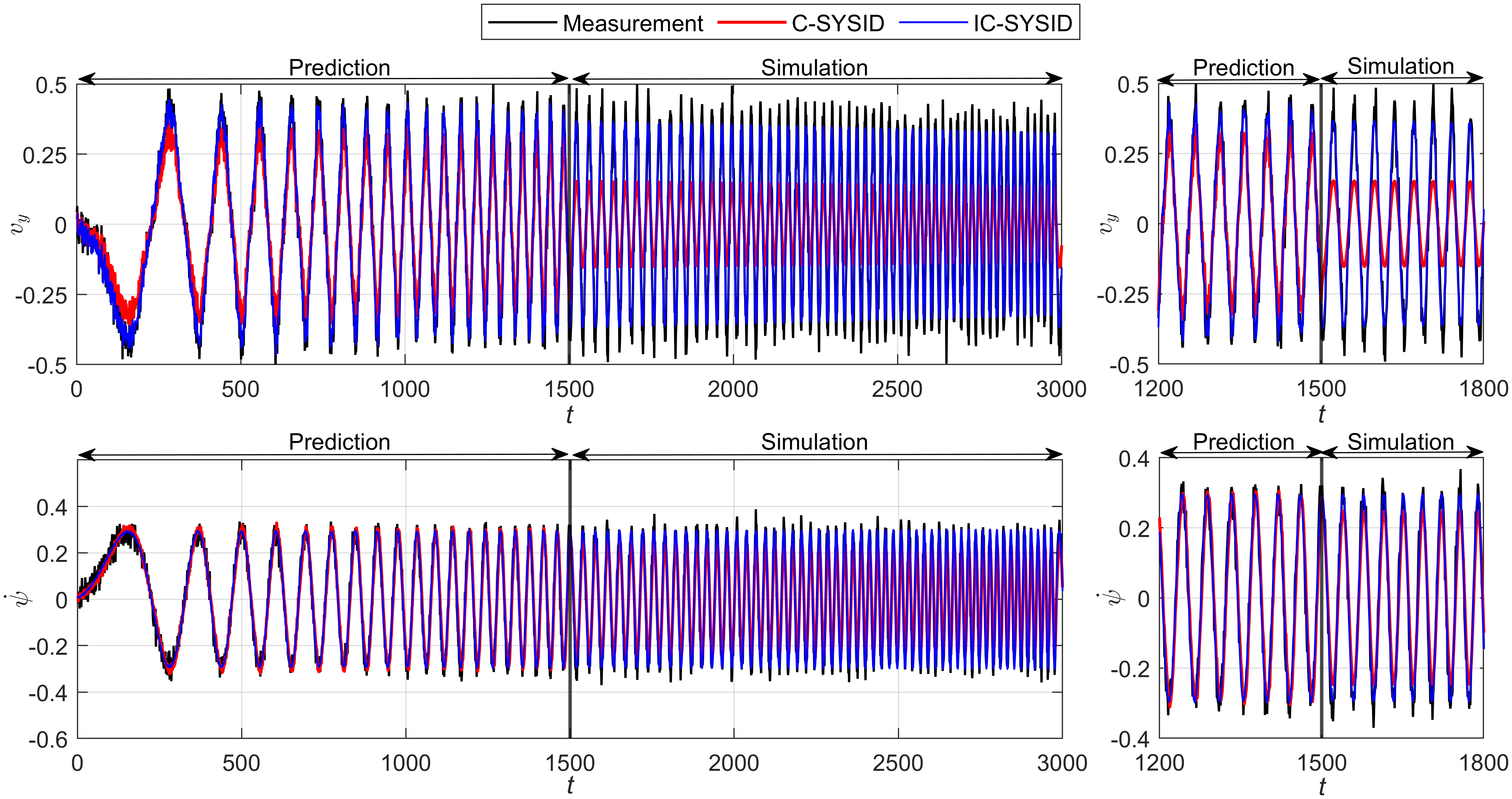}
    \caption{Comparison of C-SYSID with $K=4$ and IC-SYSID with eCC on Car Dataset (full-scale plot on the left, zoomed-in plot on the right)}
    \label{fig:car-data-comp-stab}
\end{figure*}

We analyzed the performance of the learned cluster models in both prediction and simulation modes. 
\begin{itemize}
    \item In prediction mode, the inference of the model follows the formulation in \eqref{LTI}. This mode aims to utilize the learned model with historical data to predict the next states. 
    \item  In simulation mode, the inference is as follows:
\begin{equation} \label{LTIsim}
    \tilde{\boldsymbol{x}}_{t+1} = \tilde{A}\tilde{\boldsymbol{x}}_{t} + \tilde{B}\boldsymbol{u}_{t}, \quad t = 1,2, ..., T   
\end{equation}
where $\tilde{\boldsymbol{x}}_{t}$ is the estimated state. Thus, this mode validates the model under different conditions as it uses the input data and initial conditions to simulate system behavior. 
\end{itemize}
Table \ref{tab:comp3} and \ref{tab:comp4} provide the mean $\Bar{P}_n$ values with their $\pm1$ standard error alongside the mean UC\%, $R$, and $K_R$ values, over 10 experiments. In Fig. \ref{fig:car-data-comp} and \ref{fig:car-data-comp-stab}, we presented the responses of two cluster models.

\begin{table}[t]
\centering
\footnotesize
\renewcommand{\arraystretch}{1.4}
\setlength{\tabcolsep}{3pt}
\caption{Comparative Performance Analysis of IC-SYSID in Prediction Mode over 10 experiments: Car dataset}
\label{tab:comp3}
\begin{threeparttable}
\begin{tabular}{@{}llccc|ccc@{}}
\toprule
\multicolumn{2}{c|}{} 
& \multicolumn{3}{c|}{\textbf{C-SYSID}}
& \multicolumn{3}{c}{\textbf{IC-SYSID}}
\\ \midrule
\multicolumn{2}{c|}{} 
& K=2 
& K=4 
& K=8 
& CC
& CC-CM 
& eCC 
\\ \midrule

$\bar{P}_1$ 
& 
& \(0.500 (\pm 0.057)\)
& \(0.695 (\pm 0.079)\)
& \(0.618 (\pm 0.180)\)
& \(0.812 (\pm 0.003)\)
& \(0.813 (\pm 0.003)\)
& \(0.809 (\pm 0.000)\)
\\
$\bar{P}_2$ 
& 
& \(0.639 (\pm 0.090)\)
& \(0.684 (\pm 0.058)\)
& \(0.640 (\pm 0.113)\)
& \(0.830 (\pm 0.002)\)
& \(0.830 (\pm 0.002)\)
& \(0.823 (\pm 0.001)\)
\\ \midrule

$R$ 
& 
& 1000
& 1000
& 1000
& 166.2
& 166.2
& 77
\\
$K_R$ 
& 
& 2
& 4
& 8
& 26.2
& 6.8
& 3.2
\\
UC\% 
& 
& 10\%
& 15\%
& 10\%
& 0.36\%
& 0\%
& 0\%
\\ \bottomrule

\end{tabular}

\vspace{0.5em}

\begin{tablenotes}[flushleft]
\item (1) Standard deviations of $\bar{P}_n$ are given within parentheses.
\item (2) $R$ and $K_R$ values are hyperparameters for C-SYSID.
\end{tablenotes}
\end{threeparttable}
\end{table}

\begin{table}[t]
\centering
\footnotesize
\renewcommand{\arraystretch}{1.4}
\setlength{\tabcolsep}{3pt}
\caption{Comparative Performance Analysis of IC-SYSID in Simulation Mode over 10 experiments: Car dataset}
\label{tab:comp4}
\begin{threeparttable}
\begin{tabular}{@{}l|c|ccc|ccc@{}}
\toprule
\multicolumn{1}{c|}{} 
& \multirow{2}{*}{\textbf{SS}} 
& \multicolumn{3}{c|}{\textbf{C-SYSID}}
& \multicolumn{3}{c}{\textbf{IC-SYSID}}
\\ \cline{3-8}
\multicolumn{1}{c|}{} 
& 
& K=2 
& K=4 
& K=8 
& CC
& CC-CM 
& eCC 
\\ \midrule

$\bar{P}_1$ 
& 0.704
& \(-1.222 (\pm 0.791)\)
& \(0.511 (\pm 0.178)\)
& \(0.233 (\pm 0.670)\)
& \(0.763 (\pm 0.030)\)
& \(0.772 (\pm 0.005)\)
& \(0.746 (\pm 0.002)\)
\\
$\bar{P}_2$ 
& 0.699
& \(0.559 (\pm 0.156)\)
& \(0.706 (\pm 0.049)\)
& \(0.612 (\pm 0.226)\)
& \(0.821 (\pm 0.010)\)
& \(0.818 (\pm 0.001)\)
& \(0.808 (\pm 0.001)\)
\\ \bottomrule

\end{tabular}

\vspace{0.5em}

\begin{tablenotes}[flushleft]
\item (1) Standard deviations of $\bar{P}_n$ are given within parentheses.
\end{tablenotes}
\end{threeparttable}
\end{table}

The comparative results demonstrate that IC-SYSID consistently outperforms C-SYSID in both prediction and simulation modes. Additionally, the standard deviations indicate that IC-SYSID maintains consistently high performance, whereas the performance of C-SYSID varies significantly across experiments. It's important to note that since the cluster models are trained primarily in prediction mode, their evaluation in simulation mode tends to yield lower SYSID performance, as depicted in Fig. \ref{fig:car-data-comp-stab}. Conversely, the simulation mode performances of SS models are notably lower than those of IC-SYSID. Hence, we can assert that the collaborative learning of a fleet of vehicles within the FL framework has indeed resulted in a more accurate representation of vehicle dynamics.

In our experiments with vehicles at four different velocities ($v_x$), we initially anticipated identifying $Q=4$ ground-truth clusters. As expected, the SYSID performance metrics of C-SYSID with $K=4$ demonstrated improvements compared to its $K=2$ and $K=8$ counterparts, aligning with this anticipation. Yet, we made an intriguing observation: while CC-CM represented the fleet of vehicles with $K_R=6.8$ ground-truth clusters and CC with $K_R=26.2$, both surpassed the initial expectation of $Q=4$ clusters. This indicates that the complexity of vehicle dynamics at different speeds requires more than four clusters for accurate representation. Naturally, this enhanced performance came at the cost of increased training complexity, as more learnable parameters were updated. On the other hand, eCC generated an average of $K_R=3.2$ clusters, i.e., fewer than expected ground-truth cluster models. Despite resulting in a slight degradation in SYSID performance compared to CC-CM, this finding underscores an interesting insight: vehicles operating at different velocities can effectively be clustered together.

As can be seen from Fig. \ref{fig:car-data-comp}, while an unstable cluster model might accurately forecast future states based on historical data, it fails to predict the next states in the simulation mode and thus the cluster model can not be used in analyzing the fleet of vehicle dynamics. Table \ref{tab:comp3} demonstrates that C-SYSID generates at least $10\%$ unstable cluster models, whereas IC-SYSID generates $0.036\%$ unstable cluster models with CC. We also observe that the deployment of CM eliminates the unstable cluster models trained with CC, resulting from the aggregation of similar cluster models. it's notable that eCC consistently produces stable cluster models without any instances of instability. We can conclude that integrating a regularization term in \eqref{loss} effectively reduces the generation of unstable cluster models.

\section{Conclusion and Future Work} \label{conc}

This paper introduces IC-SYSID, a collaborative learning method for SYSID problems capable of learning stable ground truth cluster models without prior dataset knowledge. To prevent instability, IC-SYSID integrates a regularization term into the loss function and initializes cluster models with scaled Glorot initialization. Utilizing a mini-batch deep learning approach, it effectively handles large SYSID datasets. IC-SYSID eliminates the need for prior data knowledge by employing CC, which initiates with one cluster and dynamically generates new clusters while grouping similar workers. Additionally, we introduce CM to merge similar cluster models post-training. To enhance efficiency, eCC is developed to reduce both the number of clusters and communication rounds compared to CC. We validated the IC-SYSID on the SYSID problem of a fleet of vehicles and demonstrated that cluster models trained by IC-SYSID represent vehicle dynamics more accurately than the ones trained by C-SYSID. 

While IC-SYSID has shown effectiveness in problems targeting the learning of LTI models, it may not fully capture the nonlinear dynamics. Thus, our future work involves extending IC-SYSID to encompass the learning of nonlinear models. 

\section*{Appendix} \label{appendix}

The parameters of each cluster model used in Section \ref{synt} are as follows: 
\begin{align*}
\label{synthA}
% \begin{gathered}
A^{(1)}& =\left[\begin{array}{lll}
0.5 & 0.3 & 0.1 \\
0.0 & 0.2 & 0.0 \\
0.1 & 0.0 & 0.3
\end{array}\right] & B^{(1)} &=\left[\begin{array}{cc}
1 & 0.5 \\
0.1 & 1 \\
0.75 & 1.5
\end{array}\right] 
\\
% \end{align*}
% \begin{align*}
A^{(2)} & =\left[\begin{array}{ccc}
-0.3 & 0.0 & 0.0 \\
0.1 & 0.4 & 0.0 \\
0.2 & 0.3 & 0.5
\end{array}\right]& B^{(2)} &=\left[\begin{array}{cc}
1.5 & 0.1 \\
0.5 & 2.5 \\
0.1 & 1.5
\end{array}\right]\\
A^{(3)} &=\left[\begin{array}{lll}
-0.1 & 0.1 & 0.1 \\
0.1 & 0.15 & 0.1 \\
0.1 & 0.0 & 0.2
\end{array}\right]&  B^{(3)} &=\left[\begin{array}{cc}
0.8 & 0.1 \\
0.1 & 1.5 \\
0.4 & 0.8
\end{array}\right] \\
A^{(4)} &=\left[\begin{array}{ccc}
0.7 & 0.25 & 0.0 \\
0.1 & 0.7 & 0.1 \\
0.0 & 0.25 & 0.7
\end{array}\right]& B^{(4)} &=\left[\begin{array}{cc}
1.0 & 0.5 \\
0.5 & 2.5 \\
0.1 & 0.8
\end{array}\right] \\
A^{(5)} &=\left[\begin{array}{ccc}
0.5 & 0.1 & 0.0 \\
0.1 & -0.2 & -0.1 \\
0.0 & -0.1 & -0.3
\end{array}\right]& B^{(5)} &=\left[\begin{array}{ll}
1.0 & 0.4 \\
0.25 & 1.5 \\
0.75 & 0.1
\end{array}\right] 
% \end{gathered}
\end{align*}

The deployed hyperparameters throughout the paper for SYSID methods are presented in Table \ref{tab:hype}. 

\begin{table}[ht]
\centering
\footnotesize
\renewcommand{\arraystretch}{1.3}
\setlength{\tabcolsep}{4pt}
\caption{Hyperparameter Configuration}
\label{tab:hype}
\begin{threeparttable}
\begin{tabular}{@{}lcccccccccc@{}}
\toprule
\textbf{Dataset} 
& \(\alpha\) 
& \(R\) 
& \textit{mbs} 
& \(\beta_1\) 
& \(\beta_2\) 
& \(\mu\) 
& \(\epsilon_{\Delta}\) 
& \(\epsilon_p\) 
& \(\epsilon_{\theta}\) 
& \(\epsilon_s\) 
\\ \midrule

Synthetic 
& 0.001 
& 1000 
& 128 
& 0.9 
& 0.999 
& 0.001 
& 0.0005 
& 0.500 
& 0.500 
& 0.100 
\\
Car 
& 0.001 
& 1000 
& 64 
& 0.9 
& 0.999 
& 0.001 
& 0.0005 
& 0.750 
& 0.250 
& 0.050 
\\ \bottomrule
\end{tabular}
\end{threeparttable}
\end{table}

\section*{Acknowledgment}
During the preparation of this work the author(s) used ChatGPT in order to refine the grammar and enhance the English language expressions. After using this tool/service, the author(s) reviewed and edited the content as needed and take(s) full responsibility for the content of the publication.

\bibliographystyle{elsarticle-num}
\bibliography{sn-bibliography}% common bib file

\begin{thebibliography}{10}
\expandafter\ifx\csname url\endcsname\relax
  \def\url#1{\texttt{#1}}\fi
\expandafter\ifx\csname urlprefix\endcsname\relax\def\urlprefix{URL }\fi
\expandafter\ifx\csname href\endcsname\relax
  \def\href#1#2{#2} \def\path#1{#1}\fi

\bibitem{ljung2010perspectives}
L.~Ljung, Perspectives on system identification, Annual Reviews in Control 34~(1) (2010) 1--12.

\bibitem{keesman2011system}
K.~J. Keesman, System identification: an introduction, Springer Science \& Business Media, 2011.

\bibitem{schon2011system}
T.~B. Sch{\"o}n, A.~Wills, B.~Ninness, System identification of nonlinear state-space models, Automatica 47~(1) (2011) 39--49.

\bibitem{bruder2019nonlinear}
D.~Bruder, C.~D. Remy, R.~Vasudevan, Nonlinear system identification of soft robot dynamics using koopman operator theory, in: 2019 International Conference on Robotics and Automation, 2019, pp. 6244--6250.

\bibitem{chiuso2019system}
A.~Chiuso, G.~Pillonetto, System identification: A machine learning perspective, Annual Review of Control, Robotics, and Autonomous Systems 2 (2019) 281--304.

\bibitem{tuna2022deep}
T.~Tuna, A.~Beke, T.~Kumbasar, Deep learning frameworks to learn prediction and simulation focused control system models, Applied Intelligence 52~(1) (2022) 662--679.

\bibitem{xin2022identifying}
L.~Xin, L.~Ye, G.~Chiu, S.~Sundaram, Identifying the dynamics of a system by leveraging data from similar systems, in: American Control Conference, 2022, pp. 818--824.

\bibitem{zhang2023multi}
T.~T. Zhang, K.~Kang, B.~D. Lee, C.~Tomlin, S.~Levine, S.~Tu, N.~Matni, Multi-task imitation learning for linear dynamical systems, in: Learning for Dynamics and Control Conference, 2023, pp. 586--599.

\bibitem{wang2023fedsysid}
H.~Wang, L.~F. Toso, J.~Anderson, Fedsysid: A federated approach to sample-efficient system identification, in: Learning for Dynamics and Control Conference, 2023, pp. 1308--1320.

\bibitem{papusha2014collaborative}
I.~Papusha, E.~Lavretsky, R.~M. Murray, Collaborative system identification via parameter consensus, in: 2014 American Control Conference, 2014, pp. 13--19.

\bibitem{mcmahan2017communication}
B.~McMahan, E.~Moore, D.~Ramage, S.~Hampson, B.~A. y~Arcas, Communication-efficient learning of deep networks from decentralized data, in: Artificial intelligence and statistics, 2017, pp. 1273--1282.

\bibitem{le2021federated}
J.~Le, X.~Lei, N.~Mu, H.~Zhang, K.~Zeng, X.~Liao, Federated continuous learning with broad network architecture, IEEE Transactions on Cybernetics 51~(8) (2021) 3874--3888.

\bibitem{kaheni2024selective}
M.~Kaheni, M.~Lippi, A.~Gasparri, M.~Franceschelli, Selective trimmed average: A resilient federated learning algorithm with deterministic guarantees on the optimality approximation, IEEE Transactions on Cybernetics (2024).

\bibitem{zhao2018federated}
Y.~Zhao, M.~Li, L.~Lai, N.~Suda, D.~Civin, V.~Chandra, Federated learning with non-iid data, arXiv:1806.00582 (2018).

\bibitem{li2020federated}
T.~Li, A.~K. Sahu, M.~Zaheer, M.~Sanjabi, A.~Talwalkar, V.~Smith, Federated optimization in heterogeneous networks, in: Proceedings of Machine Learning and Systems, Vol.~2, 2020, pp. 429--450.

\bibitem{li2023fedlga}
X.~Li, Z.~Qu, B.~Tang, Z.~Lu, Fedlga: Toward system-heterogeneity of federated learning via local gradient approximation, IEEE Transactions on Cybernetics (2023).

\bibitem{yeganeh2020inverse}
Y.~Yeganeh, A.~Farshad, N.~Navab, S.~Albarqouni, Inverse distance aggregation for federated learning with non-iid data, in: Domain Adaptation and Representation Transfer, and Distributed and Collaborative Learning, 2020, pp. 150--159.

\bibitem{zhang2022fine}
L.~Zhang, L.~Shen, L.~Ding, D.~Tao, L.-Y. Duan, Fine-tuning global model via data-free knowledge distillation for non-iid federated learning, in: Proceedings of the IEEE/CVF Conference on Computer Vision and Pattern Recognition, 2022, pp. 10174--10183.

\bibitem{ghosh2020efficient}
A.~Ghosh, J.~Chung, D.~Yin, K.~Ramchandran, An efficient framework for clustered federated learning, Advances in Neural Information Processing Systems 33 (2020) 19586--19597.

\bibitem{mansour2020three}
Y.~Mansour, M.~Mohri, J.~Ro, A.~T. Suresh, Three approaches for personalization with applications to federated learning, arXiv:2002.10619 (2020).

\bibitem{li2021federated}
C.~Li, G.~Li, P.~K. Varshney, Federated learning with soft clustering, IEEE Internet of Things Journal 9~(10) (2021) 7773--7782.

\bibitem{ruan2022fedsoft}
Y.~Ruan, C.~Joe-Wong, Fedsoft: Soft clustered federated learning with proximal local updating, Proceedings of the AAAI Conference on Artificial Intelligence 36~(7) (2022) 8124--8131.

\bibitem{toso2023learning}
L.~F. Toso, H.~Wang, J.~Anderson, Learning personalized models with clustered system identification, arXiv:2304.01395 (2023).

\bibitem{kececi2024novel}
E.~Keçeci, M.~Güzelkaya, T.~Kumbasar, A novel federated learning framework for system identification, in: 2024 8th International Artificial Intelligence and Data Processing Symposium (IDAP), 2024, pp. 1--6.

\end{thebibliography}


% %% For numbered reference style

%   \textit{\LaTeX: a document preparation system},
%   Addison Wesley, Massachusetts,

% \end{thebibliography}
\end{document}